\documentclass[conference]{IEEEtran}
\IEEEoverridecommandlockouts

\usepackage{cite}
\usepackage{amsmath,amssymb,amsfonts}
\usepackage{algorithmic}
\usepackage{graphicx}
\usepackage{textcomp}
\usepackage{epstopdf}
\usepackage{xcolor}
 \usepackage{float} 
\usepackage{multirow}
\usepackage{array}
\usepackage{caption}
\usepackage{subcaption}
\newcolumntype{L}[1]{>{\raggedright\let\newline\\\arraybackslash\hspace{0pt}}m{#1}}
\newcolumntype{C}[1]{>{\centering\let\newline\\\arraybackslash\hspace{0pt}}m{#1}}
\newcolumntype{R}[1]{>{\raggedleft\let\newline\\\arraybackslash\hspace{0pt}}m{#1}}
\usepackage[acronym,nomain]{glossaries} 

\newacronym{CNN}{cnn}{Convolution Neural Network}
\newacronym{ad}{AD}{Autonomous Driving}
\newacronym{adas}{ADAS}{Advanced Driver Assistance Systems}
\newacronym{gpu}{GPU}{Graphics Processing Unit}
\newacronym{cpu}{CPU}{Central Processing Unit}
\newacronym{hw}{HW}{Hardware}
\newacronym{sw}{SW}{Software}
\newacronym{uphf}{UPHF}{Université Polytechnique Hauts-de-France}
\newacronym{its}{ITS}{Intelligent Transport System}
\newacronym{ml}{ML}{Machine Learning}
\newacronym{dl}{DL}{Deep Learning}
\newacronym{ai}{AI}{Artificial Intelligence}

\def\BibTeX{{\rm B\kern-.05em{\sc i\kern-.025em b}\kern-.08em
    T\kern-.1667em\lower.7ex\hbox{E}\kern-.125emX}}
    
\usepackage[font=footnotesize,figurewithin=none]{caption}
\usepackage{hyperref}

\usepackage{soul}

\begin{document}

\title{Performance Prediction for Convolutional Neural Networks in Edge Devices \\}

\author{\IEEEauthorblockN{Halima Bouzidi\IEEEauthorrefmark{1}, Hamza Ouarnoughi\IEEEauthorrefmark{2}, Smail Niar\IEEEauthorrefmark{2}, Abdessamad Ait El Cadi\IEEEauthorrefmark{2}}
	\IEEEauthorblockA{\IEEEauthorrefmark{1}\'Ecole Nationale Sup\'erieure d'Informatique, Algiers, Algeria\\
					\IEEEauthorrefmark{2}Université Polytechnique Hauts-de-France, LAMIH/CNRS, Valenciennes, France\\
		Email:fh\_bouzidi@esi.dz\\
		Email: firstname.lastname@uphf.fr}
}

\maketitle

\begin{abstract}
Running Convolutional Neural Network (CNN) based applications on edge devices near the source of data can meet the latency and privacy challenges. However due to their reduced computing resources and their energy constraints, these edge devices can hardly satisfy CNN needs in processing and data storage. For these platforms, choosing the CNN with the best trade-off between accuracy and execution time while respecting Hardware constraints is crucial. 
In this paper, we present and compare five (5) of the widely used Machine Learning based methods for execution time prediction of CNNs on two (2) edge GPU platforms. For these 5 methods, we also explore
the time needed for their training and tuning their corresponding hyperparameters.
Finally, we compare times to run the prediction models on different platforms. The utilization of these methods will highly facilitate design space exploration by providing quickly the best CNN on a target edge GPU. Experimental results show that eXtreme Gradient Boosting (XGBoost) provides a less than 14.73\%  average prediction error even for unexplored and unseen CNN model architectures. Random Forest (RF) depicts comparable accuracy but needs more effort and time to be trained. The other 3 approaches (OLS, MLP and SVR) are less accurate for CNN performances estimation.
\end{abstract}

\begin{IEEEkeywords}
CNN, GPU, Performance Modeling, Multiple Linear Regression, Multi-Layer  Perceptrons, Support  Vector  Machine, Random  Forest, eXtreme Gradient Boosting.
\end{IEEEkeywords}

\section{Introduction and Motivations}
\label{Introduction}
Machine Learning (ML) approaches have recently become very popular for several use cases. 
Their fields of applications, in health, agriculture, transport, etc., are growing constantly. 
This increasing interest in ML in general and Convolution Neural Networks (\textsc{cnn}) in particular, could be explained by two main reasons.
The first one is the availability of very powerful Hardware (HW) platforms in Internet of Things (IoT), edge, fog, Cloud computing and High performance computing (HPC) platforms.
The second reason is the availability and the diversity of very large datasets on the Internet.
These datasets allow to improve the training of ML algorithms and therefore enhance their accuracy.
    
Hence, the use of \textsc{cnn} has made it possible to obtain significant improvements in terms of accuracy and flexibility. 
However, their development and the choice of the best \textsc{cnn} for a given problem is difficult. 
Each month, a large number of complex and accurate \textsc{cnn} models are proposed.
Consequently, finding the \textsc{cnn} which gives the best trade-off between accuracy and execution time is a tedious task \cite{Chen}.

For edge devices with limited computing power and real-time constraints such IoT or autonomous cars, choosing the best \textsc{cnn} implementation is problematic.
In \cite{bianco}, authors have shown that \textsc{cnn} is not necessarily correlated with the number of FLoating-point OPerations (FLOPs) or the size of the \textsc{cnn}.

Having a design tool for rapid \textsc{cnn} performance estimation becomes crucial to reduce design time and to obtain a high-performance system. 
Early execution time estimation allows to quickly determine the best \textsc{cnn} implementation for a given application requirements and a given hardware (HW) constraints. 
Depending on the \textsc{cnn} to execute, the user may need different HW to minimize inference time.
Such a tool can be used either to choose the best \textsc{cnn} for a given HW platform  and/or to explore different HW platforms for
a specific \textsc{cnn}. 

In this paper, we focus on proposing a methodology to help the designer in choosing the most efficient \textsc{cnn} for a given HW platform. 
The \textsc{cnn} execution time estimator must provide a high accuracy, a low design time and a high level of flexibility to be adapted to different \textsc{cnn}s and HW platforms. In addition, the number of estimator hyperparameters must be reduced and easy to tune. 

In this work, we compare some of the most successful state-of-the-art prediction algorithms for estimating the execution time of the \textsc{cnn} inference phase.
As these algorithms have a set of (hyper)parameters, in the rest of the paper, they are called {\it{Models}}. The following ML-based models have been considered: Multiple Linear Regression using the Ordinary Least Squares (OLS), Multi-Layer Perceptrons (MLP), Support Vector Regression (SVR), Random Forest (RF), and eXtreme Gradient Boosting (XGBoost).
Our work is not intended to propose new ML-based models but to compare the 5 models in order to analyze their strengths and weaknesses. For each of the studied models, we analyze prediction accuracy, time to tune and train the model and finally, time  to  run  the  prediction  models  on different  Hardware platforms.  

The remainder of this paper is organized as follows.
Section \ref{RelatedWorks} gives a literature review of \textsc{cnn} performance estimation and design space exploration. 
In section \ref{Approach}, we first analyze and discuss the \textsc{cnn} features that have to be considered in the execution time prediction, then a survey of the used ML-based approaches is presented. 
Section \ref{Evaluation} is devoted to experimental results. 
Finally, conclusion and future work will be given in section \ref{conclusion}.

\section{Related works}\label{RelatedWorks}
Motivated by speed and security purposes, there was recently a trend towards migrating ML applications from cloud and central computing to edge computing. 
To find a good matching between \textsc{cnn} requirements in terms of speed and energy consumption either in the training or in the inference phases, several projects have been conducted recently targeting edge platforms.

In \cite{comparaisonGPU}, the authors compare: linear regression, support vector machines and random forests with a 
Bulk synchronous parallel (BSP) based analytical model. Machine learning approaches have been used to provide reasonable predictions without detailed knowledge of application code or hardware characteristics. The authors use profiling information from 9 benchmarks on 9 different \textsc{gpu}s. In opposite to our work, their method is not dedicated to \textsc{ML} applications. In addition, in our work we use a larger set of benchmarks and we obtain a better accuracy.  

In \cite{PredictingComputational}, the authors propose a methodology to estimate training time for \textsc{cnn}s. In their approach, training time is seen as the product of the training time per epoch and the number of epochs which need to be performed to reach the desired level of accuracy.  The deep learning network is decomposed in several parts and the execution time estimation operates on these parts.  Timings for these individual parts are then combined to provide a prediction for the whole \textsc{cnn} execution time. 

In \cite{alibaba}, the authors propose an analytical framework to characterize deep learning training workloads in large AI clouds and clusters.  Using different training architectures, the authors try to identify performance bottleneck for various DNN workloads.  For the training phase, weight and gradient communication consumes almost 62\% of the total execution time on average. Their simple analytical performance model is based on the key workload features and allows only to identify architectural bottlenecks. 

PALEO \cite{paleo} is another framework for estimating training execution time. In PALEO the number of floating point operations required for an epoch is  multiplied  by a scaling factor to obtain execution time for the training phase. However, PALEO does not take into account numerous other operations which do not scale linearly with the number of floating point operations and which has a big impact on execution time. 
In the literature we may also find several tools dedicated to power and memory optimization for edge HW platforms.

In \cite{fineGrainedEnergy}, the authors proposed a multi-variable linear regression model to predict energy consumption of \textsc{cnn}s based on the number of SIMD instructions and main memory accesses. They used an Nvidia Jetson TX1 \textsc{gpu} and they obtained an average relative error of about 20\%. Our profiling and analysis phases, detailed in Section \ref{Approach}, have shown that these 2 parameters must be enhanced by others to obtain a more accurate estimation.

The idea behind tools in \cite{NeuralPower, Hyperpower} is to explore the  hyperparameter space for NNs running on a given hardware platform and to estimate power and memory usages. In most of the existing projects, a learning-based polynomial regression approach is used.
In our work, we do not treat power consumption estimation. Only  tools for execution time estimation are presented and compared. However, as we will see in the following sections, the same approaches can be developed for power and energy consumption.

\section{Proposed Approach}
\label{Approach}
\begin{figure*}
\centering
\includegraphics[width=0.8\textwidth, height=0.18\textheight]{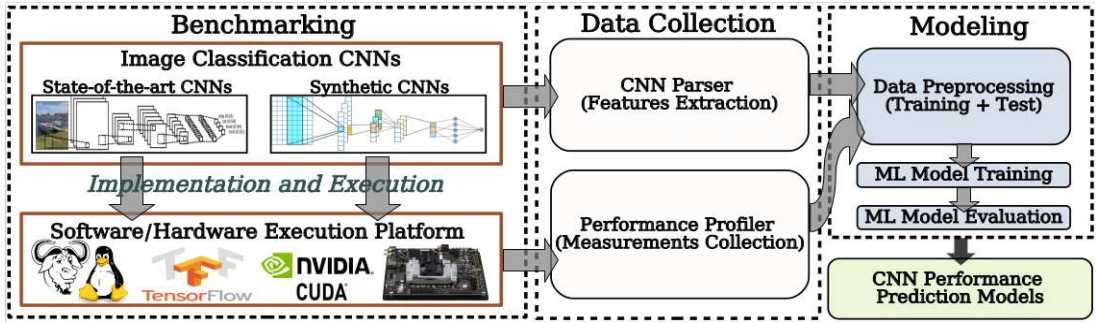} 
\caption{Modeling methodology for \textsc{cnn}s execution time prediction models.}
\label{fig:overview}
\end{figure*}

Figure \ref{fig:overview} gives an overview of our proposed modeling methodology. The approach includes three main steps. The benchmarking step aims to define the benchmarks characteristics and determine the most impacting \textsc{cnn} features. To make the model simple, only these features will be taken into account in the models. 
In the data collection step, we extract the data used in the elaboration of the \textsc{cnn} execution time prediction.
Finally, the modeling step details the used ML-models, their training process with hyperparameters tuning and their evaluation. 
 As explained in the following section, each of the considered models includes a set of {\it{hyperparameters}} and a set of {\it{parameters}}. The hyperparameters are different from one model to another. They draw the model architecture. The parameters produce one instance of the obtained model and explain the manner by which the model is implemented for the prediction.
As example for MLP, the hyperparameters correspond to the number of neurons, the number of hidden layers, the activation function, etc. The parameters for MLP correspond to the weights and the biases values used in the model. More details will be given in sub-Section \ref{sub:ML-approaches}.

\subsection{Problem Formulation}
Predicting \textsc{cnn}s inference time on \textsc{gpu} platforms can be formulated as follows: 
As inputs, we have a \textsc{cnn} (noted $CNN_i$) characterized by a set of $n$ features ${f_1, f_2, \dots, f_n}$, and a \textsc{gpu} platform (noted $GPU_j$) represented by a set of $m$ Hardware characteristics ${c_1, c_2, \dots, c_m}$. The number of convolutional and fully-connected layers, the image size, and the number of neurons are examples of \textsc{cnn} features. Platform parameters may correspond to the number of cores in the \textsc{gpu}, the memory size, the clock frequency, etc.  

An inference time prediction function (noted $T$) is a mapping function from $CNN_i$ and $GPU_j$ to $ \mathbb {R_{+}}$
\begin{center}
$T: CNN_i,GPU_j \xrightarrow{} \hat{y}$
\end{center}
\begin{equation}
\hat{y} = T(f_1, f_2,\dots, f_n, c_1, c_2, \dots, c_m)
\label{eq:regression}
\end{equation} 
where $\hat{y}$ is the predicted inference time of $CNN_i$ on the $GPU_j$.

In this paper, due to the lack of space, we only present the case of two different edge \textsc{gpu}s, namely Nvidia Jetson AGX Xavier\cite{AGX} and Nvidia Jetson TX2 \cite{TX2}. 

The same modeling methodology has been used on each edge \textsc{gpu} platform and two sets of models have been obtained.
The equation \ref{eq:regression} becomes:
\begin{equation}
\hat{y} = T_{GPU_j}(f_1, f_2,\dots, f_n)
\label{eq:regression_2}
\end{equation}
where $GPU_j \in \{AGX, TX2\}$. 

In this context, our approach aims to answer the following questions: 
\begin{enumerate}
\item What are the most important \textsc{cnn} features that impact inference time? Our answer is given in sub-section \ref{sub:features}.
\item What are the most relevant modeling algorithms that give the best accurate predictions of \textsc{cnn}s inference time? Our answer is given in sub-section \ref{sub:ML-approaches}.
\end{enumerate}

Our modeling approach is based on finding the relationship between the input \textsc{cnn} features and observed inference time.
For this purpose, we rely on regression models which are part of supervised ML algorithms.   
    
\begin{figure*}
    \centering
    \begin{minipage}{0.42\textwidth}
    \centering
    \captionsetup{justification=centering}
    \captionof{table}{\textsc{FLOPs estimated by TensorFlow Profiler of some state of the art \textsc{cnn}s. (\textbf{B:} Billions, and \textbf{M:}  Millions).}}
    \label{tab:table-flops}
    \small
    \begin{tabular}{| c | c | c | c | c | c |}
    \hline
    CNN Model       & Conv2D & Add    & Mul    & Pooling \\ \hline
    ResNet-50       & 7.71B   & 31.02M & 25.58M & 1.81M   \\ \hline
    DenseNet-121    & 5.67B   & 7.89M  & 8.02M  & 1.98M   \\ \hline
    DPN-98          & 23.34B & 70.54M & 61.63M & 2.71M   \\ \hline
    GoogleNet       & 3.00B  & 6.61M  & 6.64M  & 12.55M  \\ \hline
    ResNet-101V2       & 14.38B  & 52.32M  & 44.59M  & 2.16M  \\ \hline
    Inception-v3       & 5.67B  & 23.80M  & 23.85M  & 12.18M  \\ \hline
    \end{tabular}
    
    \end{minipage}
    \begin{minipage}{0.5\textwidth}
        \centering
    \includegraphics[width=0.9\textwidth, height=0.22\textheight]{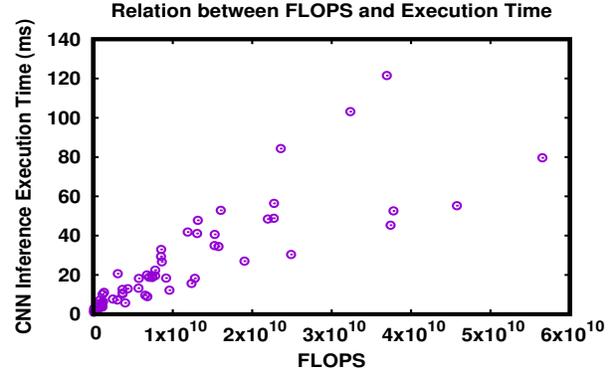}    
    \caption{Execution time versus floating-point operations (FLOPs). This figure demonstrates that using only the numbers of FLOPS is not sufficient to estimate execution time. This figure gives the execution time of some of our \textsc{cnn} benchmarks detailed in Section IV ranked by their FLOPs.}
    \label{fig:Infer_flops_224}
        \end{minipage}
    \end{figure*}

\subsection{CNN Features for our Prediction Models}
\label{sub:features}
\textsc{cnn}s inference time is mainly impacted by the following factors: 

\begin{enumerate}
\item Computational complexity, which corresponds to \textsc{gpu} cores activities; 
\item Memory requirements, which corresponds to read and write memory operations;
\item \textsc{cnn} internal properties, which corresponds to dependencies between computation operations and memory operations.
\end{enumerate}
These features  are detailed below. \\

\subsubsection{Computational complexity}
The total number of FLOPs is  used to measure the computational complexity when performing \textsc{cnn}s inference. 
As reported in table \ref{tab:table-flops}, this number is highly dominated by the number of operations performed in convolutional layers. 
Convolutional layers represent the bottlenecks of computations in \textsc{cnn}s. 
Our experimentation shown in figure \ref{fig:Infer_flops_224} confirms that \textsc{cnn}s inference time is not linearly correlated to the number of FLOPs. 
We can then conclude that considering only the computational complexity is not enough to predict accurately the \textsc{cnn} inference time.\\
    
\subsubsection{Memory requirements}
Memory activities have a significant impact on the execution time on \textsc{gpu}s. 
However, it's hard to extract the information about memory activities and requirements without profiling the \textsc{cnn}s on the \textsc{gpu} platform.
This solution must be avoided during the prediction as it's a time consuming task and will also add a significant overhead to the prediction latency. 
To overcome this problem, we rely on some specific characteristics of \textsc{cnn}s that are correlated to memory activities. 
Our experiments show that the major memory requirements of \textsc{cnn}s can be devoted mainly to 3 factors: 
\begin{enumerate}
    \item Reading \textsc{cnn}s parameters, i.e weights and biases,
    \item Reading the input data, writing the output results and 
    \item Reading and writing the intermediate data of the hidden layers, i.e activations. 
\end{enumerate}
During the \textsc{cnn} inference, convolutional filters and activations are constantly accessed  which increases the inference time \cite{8573527}. 
Moreover, this time increases when activations and weights can not be mapped entirely in cache memories. Cache misses considerably increase the \textsc{cnn} inference time. 
Given the above facts, we assume that both weights and activations are highly impacting memory activities when performing \textsc{cnn}s inference. \\

\subsubsection{CNN internal properties} 
In addition to computational complexity and memory requirements, other properties related to \textsc{cnn}s internal architecture impacting inference time have to be taken into account. 
We have also considered the \textbf{Weighted sum of neurons} in  convolutional layers. 
This metric is calculated by multiplying the number of neurons in convolutional layers by the filter size: $height \times width \times depth$. 
The idea here is to give more importance to neurons with large filter sizes.
The number of neurons in fully connected layers is taken as it is, because the neuron is associated to a single weight. 

\begin{figure}[ht!]
    \centering
    \includegraphics[width=0.49\textwidth, height=0.25\textwidth]{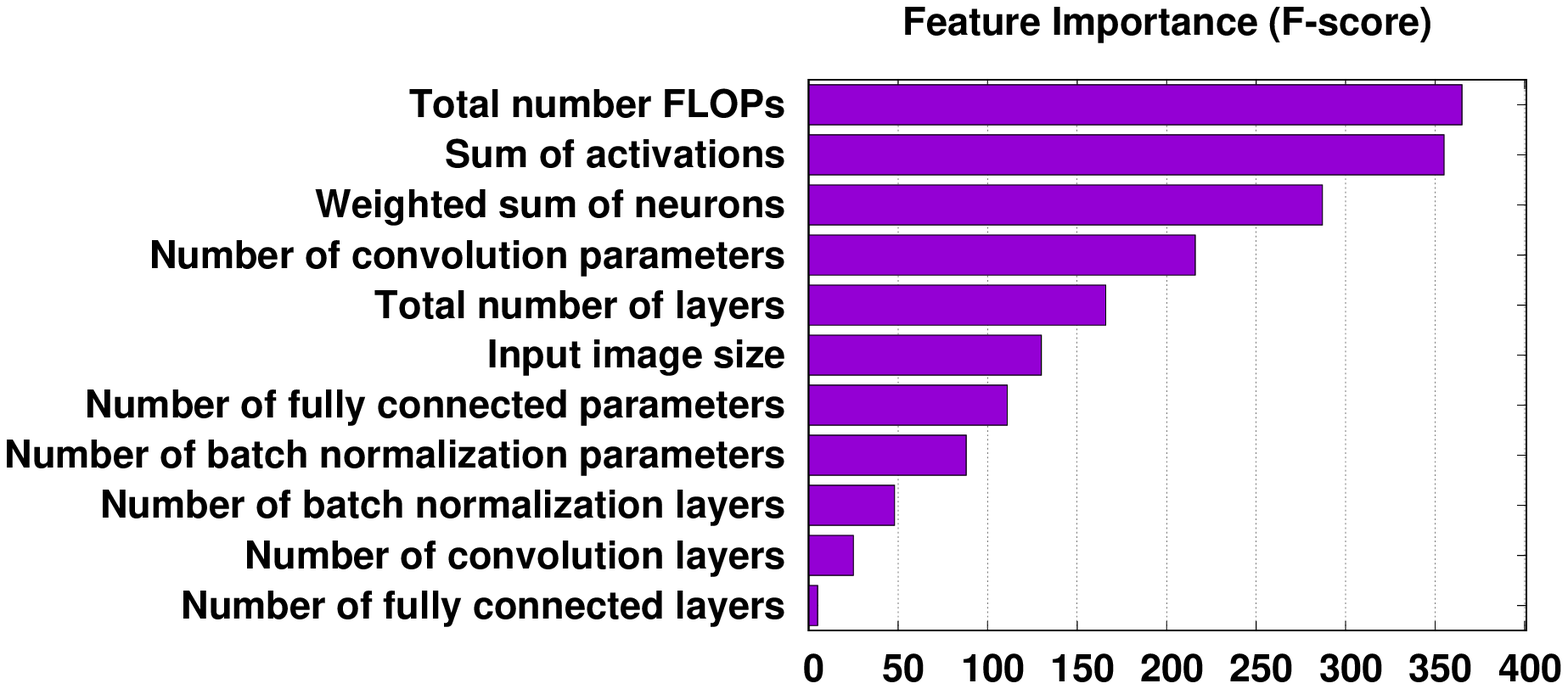}
    \caption{The feature importance calculated by XGBoost. Features with a high F-score are the most impacting on the \textsc{cnn}s inference time.}
    \label{fig:feature_importance_xgboost}
\end{figure}

To select the most important \textsc{cnn}'s features in the execution time prediction, we used the F-score metric of the XGBoost algorithm. This technique is used for all of the prediction models detailed in section \ref{sub:ML-approaches}.
The considered features and their corresponding impacts are depicted in figure \ref{fig:feature_importance_xgboost}. \\

\begin{table*}[ht!]
\centering
\captionsetup{justification=centering}
    \captionof{table}{\textsc{Evolution of the adjusted $R^{2}$ during the stepwise linear regression. Features are added to the OLS prediction model one by one, from left (most important feature) to right (less important feature) of the table}}
    \label{tab:table-stepwise-lr}
\begin{tabular}{|c|p{0.06\textwidth}|p{0.06\textwidth}|p{0.06\textwidth}|p{0.06\textwidth}|p{0.06\textwidth}|p{0.06\textwidth}|p{0.06\textwidth}|p{0.06\textwidth}|p{0.06\textwidth}|p{0.06\textwidth}|p{0.06\textwidth}|}
\hline
\textbf{Features} & Total number of FLOPs & Sum of activations & Weighted sum of neurons & Number of convolution parameters & Total number of layers & Input image size & Number of fully connected parameters & Number of batch normalization parameters & Number of batch normalization layers & Number of convolutional layers &  Number of fully connected layers \\
\hline
\textbf{Adjusted $R^{2}$} & 0.970 & 0.985 & 0.987 & 0.988 & 0.988 & 0.988 & 0.988 & 0.988 & 0.988 & 0.988 & 0.988 \\ 
\hline
\end{tabular}
\end{table*}

\subsection{Modeling Approaches}
\label{sub:ML-approaches}
Five ML-based algorithms have been used to design the \textsc{cnn} inference time prediction models:
\begin{enumerate}
\item Multiple Linear Regression using the Ordinary Least Squares (OLS)
\item Multi-Layer Perceptrons (MLP)
\item Support Vector Regression (SVR)
\item Random Forest (RF)
\item eXtreme Gradient Boosting (XGBoost)
\end{enumerate}

In order to tune each model's hyperparameters, we applied an exhaustive grid search.
During this process, we perform a K-fold cross validation \cite{rodriguez2009sensitivity} to select the best hyperparameters values and combination.
Once the hyperparameters are tuned, the training of the chosen ML algorithms is performed on the parameters using the training dataset. 
Finally, the obtained prediction models are evaluated using the test dataset.
We give an overview of the ML algorithms used in our approach in the following. \\

\subsubsection{Multiple Linear Regression using the Ordinary Least Squares (OLS)}
Multiple linear regression approaches use a linear function to model the relationship between a target variable ($y$) and $n$ input features ($X_1, X_2, \dots, X_n$):
\begin{equation}
y = \alpha _nX_n+ \alpha _{n-1}X_{n-1}... +\alpha _2X_2 + \alpha _1X_1 +\beta + \epsilon
\end{equation} 
In this equation  $\alpha 1, \alpha 2, \dots, \alpha_n$  and $\beta$ are the parameters of the MLR. They are fixed during the training phase in such a way to minimize the difference ($\epsilon$) between estimated ($\hat{y}$) and real ($y$) values of the inference execution times. 

Among the variety of Multiple Linear Regression algorithms, we have chosen the most frequently used Ordinary Least Squares regression (OLS) \cite{matloff2017statistical}.   
In our study, we have applied a stepwise regression to choose the best combination of features with the highest impact on \textsc{cnn}s inference time. The results of the stepwise regression are given in table \ref{tab:table-stepwise-lr}.
The features are added based on their importance showed in the figure \ref{fig:feature_importance_xgboost}.

We notice that the model has achieved a highest adjusted $R^{2}$ (98.8\%) with only four input features, namely : FLOPs, sum of activations, weighted sum of neurons, and number of convolutional layers parameters. 
The reasons behind choosing Multiple Linear Regression for \textsc{cnn}s inference time prediction are: 1) their simplicity in the implementation, and 2) their short training time. However, this approach is not accurate when the linear relationship between input features and the target variable, execution time in our case,  is not valid. 
To overcome this limitation, other algorithms which have the ability to model both linear and non-linear relationships have been implemented and compared. These algorithms are presented in the following sub-sections. \\

\subsubsection{Multi-Layer Perceptrons}
The Multi-Layer Perceptrons (MLP) \cite{murtagh1991multilayer} is a part of Artificial Neural Networks (ANN). 
The MLP is a succession of hidden layers where each layer applies the following transformation on the input features: 

\begin{equation}
\hat{z} = \theta (\sum_{i=1}^{n}w_{i}x_{i} + b)
\end{equation}
where $\hat{z}$ is the output value,  $w_i$ is the weight of the feature $x_i$, $b$ is the bias and $\theta$ is the activation function. 
Choosing the good MLP architecture is done by fixing the hyperparameters, such as the number of layers, number of neurons, types of activation function, etc. We rely on the grid search method with the K-fold cross validation to choose the best hyperparameters values. Once these hyperparameters are fixed, parameters values of $w_{i}$ and $b$ are fixed at the training phase. \\

\subsubsection{Support Vector Regression}
Support Vector Machine \cite{hearst1998support} is one of the most powerful data driven algorithms used for classification and regression problems.
The SVR deals with non-linearity by using kernels which are functions that map the input data from the original space to a higher dimensional space where features can be linearly separable.
The most important hyperparameters for the SVR are:
\begin{itemize}
\item Kernel type: linear, polynomial, sigmoid, and radial basis function (RBF).
\item Gamma: the kernel coefficient for the polynomial, sigmoid and RBF kernels.
\item Cost (C): regularization factor which controls the trade-off between the training error and the generalization of the model.
\item Epsilon, which defines the interval of errors within which no penalty is applied to training loss function.
\end{itemize} 

These hyperparameters have been tuned using the grid search method. \\

\subsubsection{Random Forest}
Unlike the above methods based on the best line or hyperplane fitting, Random Forest (RF) uses decision trees.
RF is based on the bagging technique where predictions are made by multiple predictors (i.e. decision trees). 
Each of them is trained on a subset of training samples and a subset of features selected randomly. 
The final prediction is calculated by averaging the predictors outputs. 
We tune the following hyperparameters:
\begin{itemize}
\item Number of predictors needed to train the model,
\item Maximum depth of each decision tree,
\item Minimum number of samples required to split nodes,
\item Minimum number of samples required for a leaf node,
\item Maximum number of features to consider when splitting the nodes.
\end{itemize} 

We have also used the bootstrap method for data sampling. \\

\subsubsection{eXtreme Gradient Boosting}
eXtreme Gradient Boosting (XGBoost) \cite{chen2016xgboost} is a decision-tree based algorithm. 
Unlike  Random Forest, XGBoost is based on the boosting technique. The boosting technique makes predictions from weak predictors that are arranged sequentially: 
the first predictor is trained on the entire dataset, where the subsequent predictors are trained on the residuals of the prior predictors. 
This technique helps to focus on the mispredicted values. 
The algorithm uses the gradient descent algorithm to minimize prediction errors. 
As explained in Section \ref {sub:features}, we use first the feature importance calculated by XGBoost to select the most impacting features (see figure \ref{fig:feature_importance_xgboost}). These features are then used by the performance estimator. 
XGBoost has several hyperparameters to tune, which can be categorized into three groups: 
\begin{itemize}
\item General hyperparameters,
\item Booster hyperparameters,
\item Learning hyperparameters.
\end{itemize} 
Once the optimal combination of hyperparameters is obtained, we perform a full training on the entire training dataset, with the early stopping technique.

\section{Experimental Results}
\label{Evaluation}
This section details the followed evaluation methodology and the obtained evaluation results.
We used two Nvidia \textsc{gpu} platforms, namely the  Jetson TX2 and the Jetson AGX Xavier. 
The hardware specifications of each platform is described in the table \ref{tab:agx_tx2}. We have configured the platforms to profiling mode (maximum power mode) in order to minimize the host CPU interference.
\begin{table}[ht!]
\captionsetup{justification=centering}
    \captionof{table}{\textsc{Hardware platforms used in the experiments}}
\label{tab:agx_tx2}
\begin{tabular}{| l | C{.15\textwidth} | C{.15\textwidth} |}
\hline
\textbf{Hardware feature}	&	\textbf{Jetson TX2}			&	\textbf{Jetson AGX Xavier}		\\
\hline
CPU(ARM)					& 6-core Denver and A57 2 GHz	&	8-core Carmel 2.26 GHz	\\
\hline
Memory						& 8 GB 128-bit LPDDR4 			&	16 GB 256-bit LPDDR4x			\\
\hline
Memory bandwidth			& 58.4 GB/s						& 	137 GB/s \\
\hline
GPU							& 256-core Pascal 1.3 GHz		& 	512-core Volta 1.37 GHz + 64 tensor cores \\
\hline
Power						& 7.5W/15W 						&	10W/15W/30W \\
\hline
\end{tabular}
\end{table}  
We have used the same underlying software configuration in the two edge \textsc{gpu} platforms. 
\textsc{cnn}s have been implemented on \textsc{gpu} using the Keras 2.3.1 API with TensorFlow 1.14 as backend \cite{abadi2016tensorflow}.
This framework is running on top of Cuda version of 10.0 and a cuDNN version of 7.5.3.
The host operating system in both platforms is Linux Ubuntu 18.04.3 LTS with a kernel 4.9.149-tegra. 

\subsection{Evaluation Methodology}
\label{sub:Evaluation_Methodology}

As shown in figure \ref{fig:overview}, our modeling process is subdivided into three main steps: 1) Benchmarking, 2) Data collection and 3) Modeling.\\
 
\subsubsection{Benchmarking step}
\label{subsub:benchmark}
The benchmarking step is mainly based on profiling \textsc{cnn}s inference  on different \textsc{gpu}s. 
The experiments have been designed based on executing state-of-the-art image classification \textsc{cnn}s by varying the parameters presented in table \ref{tab:parameters}.

\begin{table}[ht!]
\centering
\captionsetup{justification=centering}
    \captionof{table}{\textsc{Details of the benchmarks used in the experiments. In total, we have 2056 and 1975 (respectively) input-data for AGX and TX2 (respectively) for 5 performance estimation models: 70\% have been used for training and 30\% for tests and accuracy calculations}}
\label{tab:parameters}
\begin{tabular}{|l|l|l|}
\hline
\begin{tabular}[c]{@{}l@{}}CNN\\ Architectures\end{tabular} & \begin{tabular}[c]{@{}l@{}} \# CNN\\ variants\end{tabular} & Input Image Sizes (squared) \\ \hline
GoogleNet & 1 & \begin{tabular}[c]{@{}l@{}}{[}224,240,256,299,320,331,448,480,\\ 512,568,600,720,800,896,1024{]}\end{tabular} \\ \hline
Inception V3 & 1 & \multirow{3}{*}{\begin{tabular}[c]{@{}l@{}}{[}75,90,112,128,150,224,240,256,\\ 299,320,331,448,480,512,568,600,\\ 720,800,896,1024{]}\end{tabular}} \\ \cline{1-2}
\multirow{2}{*}{InceptionResNet V2} & \multirow{2}{*}{1} &  \\
 &  &  \\ \hline
DPN & 4 & \multirow{4}{*}{\begin{tabular}[c]{@{}l@{}}{[}32,56,64,75,90,112,128,150,224,\\ 240,256,299,320,331,448,480,512,\\ 568,600,720,800,896,1024{]}\end{tabular}} \\ \cline{1-2}
DenseNet & 5 &  \\ \cline{1-2}
\multirow{2}{*}{Xception} & \multirow{2}{*}{1} &  \\
 &  &  \\ \hline
EfficientNet & 4 & \multirow{3}{*}{\begin{tabular}[c]{@{}l@{}}{[}32,56,64,75,90,112,128,150,224,\\ 240,256,299,320,331,448,480,512,\\ 568,600,720,800,896,1024,1200,1600{]}\end{tabular}} \\ \cline{1-2}
\multirow{2}{*}{MNASNet} & \multirow{2}{*}{5} &  \\
 &  &  \\ \hline
ResNet V1 & 12 & \multirow{9}{*}{\begin{tabular}[c]{@{}l@{}}{[}32,56,64,75,90,112,128,150,224,\\ 240,256,299,320,331,448,480,512,\\ 568,600,720,800,896,1024,\\ 1200,1600,1792,2400{]}\end{tabular}} \\ \cline{1-2}
ResNet V2 & 7 &  \\ \cline{1-2}
MobileNet V1 & 4 &  \\ \cline{1-2}
MobileNet V2 & 5 &  \\ \cline{1-2}
MobileNet V3 & 4 &  \\ \cline{1-2}
ResNext & 2 &  \\ \cline{1-2}
SENet & 6 &  \\ \cline{1-2}
ShuffleNet V1 & 4 &  \\ \cline{1-2}
ShuffleNet V2 & 3 &  \\ \hline
\end{tabular}%
\end{table}

As shown in table \ref{tab:parameters}, we have implemented the state-of-the-art \textsc{cnn}s dedicated to image classification.  \textsc{cnn} architectures such as: Inception, ResNet, MobileNet, etc., have been tested on the two \textsc{gpu} platforms. An FP-32 bits representation for the weights and tensors has been used. The weights of these \textsc{cnn}s have been set randomly as their values have no impact on the inference time. We varied three factors: 
\begin{enumerate}
\item {Input Image Sizes}: Here the impact of the image size on the inference time is studied. We tested the frequently used image sizes in the literature from 32*32 to 2400*2400 pixels depending on the \textsc{cnn}. 
\item {\textsc{cnn} Variants}: Different variants of the same \textsc{cnn} architecture are considered in order to extend our dataset. Some of the considered \textsc{cnn}, such as ResNet V1, have up to 12 different variants.
\item {\textsc{cnn} Architectures}: Finally, we consider different architectures of \textsc{cnn} to quantify their impact on the inference time.

\end{enumerate} 
In total, we obtained a dataset of 2056 and 1975 (respectively) on AGX and TX2 (respectively) as inputs for our performance estimation models. This difference is due to the fact that the TX2 \textsc{gpu} platform has a smaller memory and some of the benchmarks couldn't be executed. In the experiments, 70\% of the input-data has been used for training and 30\% for tests and accuracy measurements. In the experimental results section, we will evaluate the accuracy of each of the 5 models, when exploration is realized in 3 groups. \\

\subsubsection{Data collection step}
The dataset used for \textsc{cnn}s inference time modeling is collected from two sources: 
\begin{itemize}
\item {\textsc{cnn}s implementation descriptions}: we have developed a parser module that takes \textsc{cnn}s implementation as input and gives their important feature values (see figure \ref{fig:feature_importance_xgboost}).
\item {Performance measurements}: \textsc{cnn}s inference times measurements have been collected using the Nvidia profiling tool Nvprof \cite{nvprof}. 
Each inference has been ran 100 times on 100 randomly chosen images in order to minimize the profiling overhead. \\
\end{itemize}

At this point the dataset is collected and ready to use in the modeling step.\\ 
   
\subsubsection{Modeling step}
\label{subsub:modeling}
\begin{figure}[!ht]
    \centering
    \includegraphics[width=0.4\textwidth, height=0.35\textwidth]{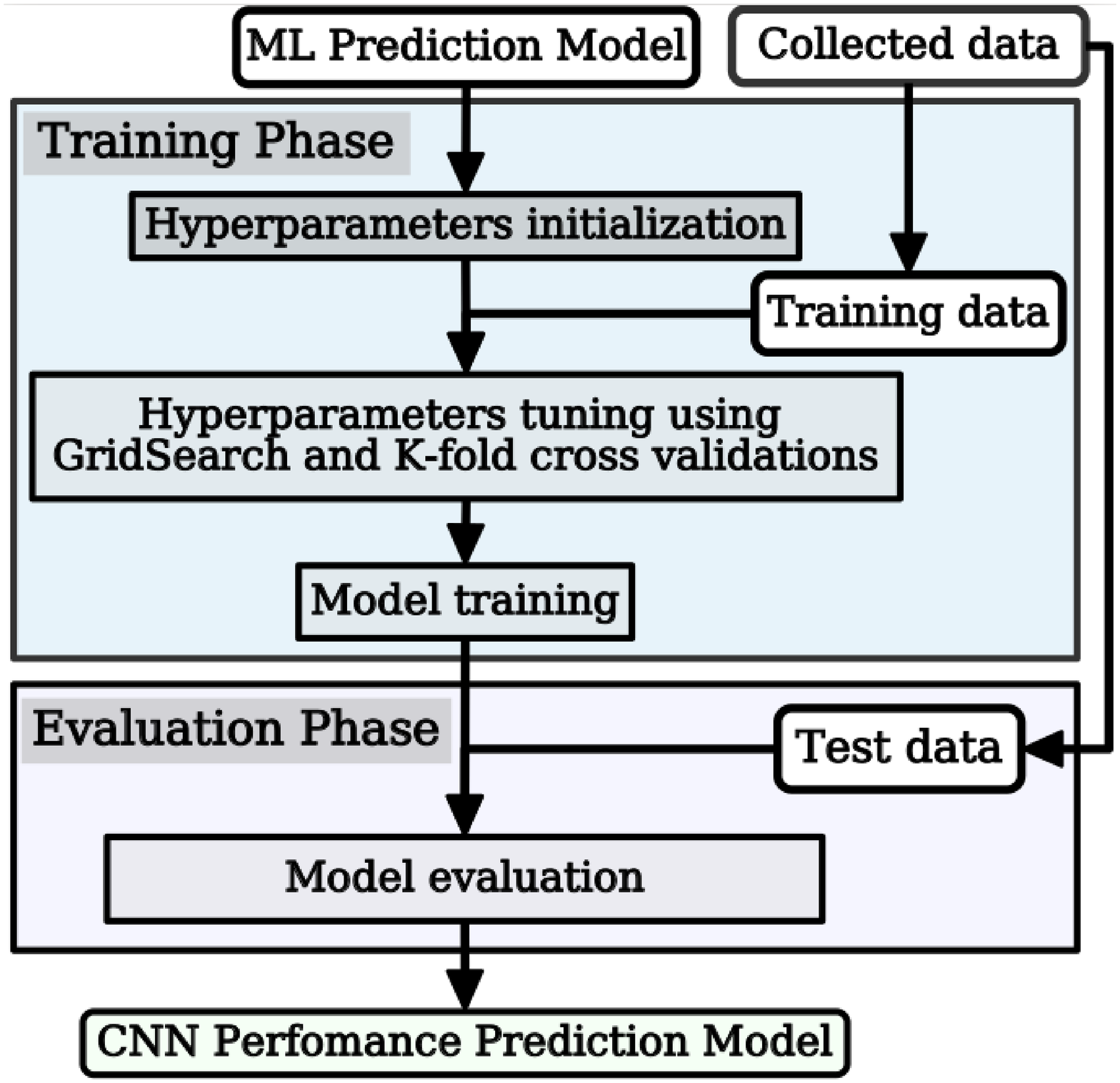}
    \caption{The process of constructing the prediction models.}
    \label{fig:ml_model_construction}
\end{figure}

Figure \ref{fig:ml_model_construction} details the modeling steps followed to obtain the 5 prediction models.

As shown in the figure \ref{fig:ml_model_construction}, our modeling process has two inputs:
\begin{enumerate}
\item \textbf{ML prediction model name}, which corresponds to one of the five ML algorithms described in section \ref{sub:ML-approaches}.
\item \textbf{Collected data}, which is the dataset obtained in the data collection step as described in Table \ref{tab:parameters}.
\end{enumerate}
The collected data is split into training data and test data. 
During the training phase, the search space for the hyperparameters is firstly initialized for each prediction model before being tuned using the training data.
The tuning is realized using grid search and K-fold cross validation techniques in order to select the best values of hyperparameters.
Finally, the prediction models are trained with the obtained optimal hyperparameters.

To evaluate the obtained models on different configurations we have also split the test data into three exploration spaces:
\begin{itemize}
    \item \textbf{Performance estimation of New Image Sizes (NIS):} In this first group of experiments, we evaluate our 5 models on state-of-the-art \textsc{cnn} models with new input sizes. 
    We varied the input size from 32*32 to 2048*2048 pixels. We obtained different values of FLOPs, sum of intermediate activations, sum of neurons and eventually different number of parameters. The number and the type of layers remain the same.
    
    \item \textbf{Performance estimation of New \textsc{cnn} Variants (NCV): } In this second group of experiments, we measure performance estimation when new variants of the same \textsc{cnn} is considered. 
   Based on an original \textsc{cnn} architecture, new \textsc{cnn} variants have been obtained by changing the features such as the number of layers and the used operators. If we consider ResNet V1 as example, we trained our predictors on ResNetV1-20 to  ResnetV1-100 and we predict inference time for ResNetV1-200. This leads to  new numbers of: FLOPs, activations, neurons,  parameters and  layers.
    
    \item \textbf{Performance estimation of New \textsc{cnn} Architectures (NCA):} In this group of experiments, we evaluate our predictors on new hand crafted synthetic \textsc{cnn} models. 
    We randomly generate synthetic \textsc{cnn} architectures where the input size, number of layers, type of layers, number of filters, type and parameters of convolutions, parameters of fully-connected layers, and type of pooling and batch normalization layers, have been randomly set. This exploration space is the most difficult as we evaluate the capacity of the 5 models to predict the performances of completely new \textsc{cnn} architectures not included  in the training dataset. The aim of this $3^{rd}$ group of experiments is to measure the capacity of the 5 models to estimate execution time of completely new and unseen \textsc{cnn} architectures only by characterizing them through their features.
\end{itemize}
 
Inference execution time accuracy is measured using the Mean Absolute Percentage Error (MAPE). 

\begin{figure*}[!ht]
    \centering
    \begin{minipage}{0.45\textwidth}
        \centering
		\label{fig:agx_bar_error}
    \includegraphics[width=1.0\textwidth, height=0.25\textheight]{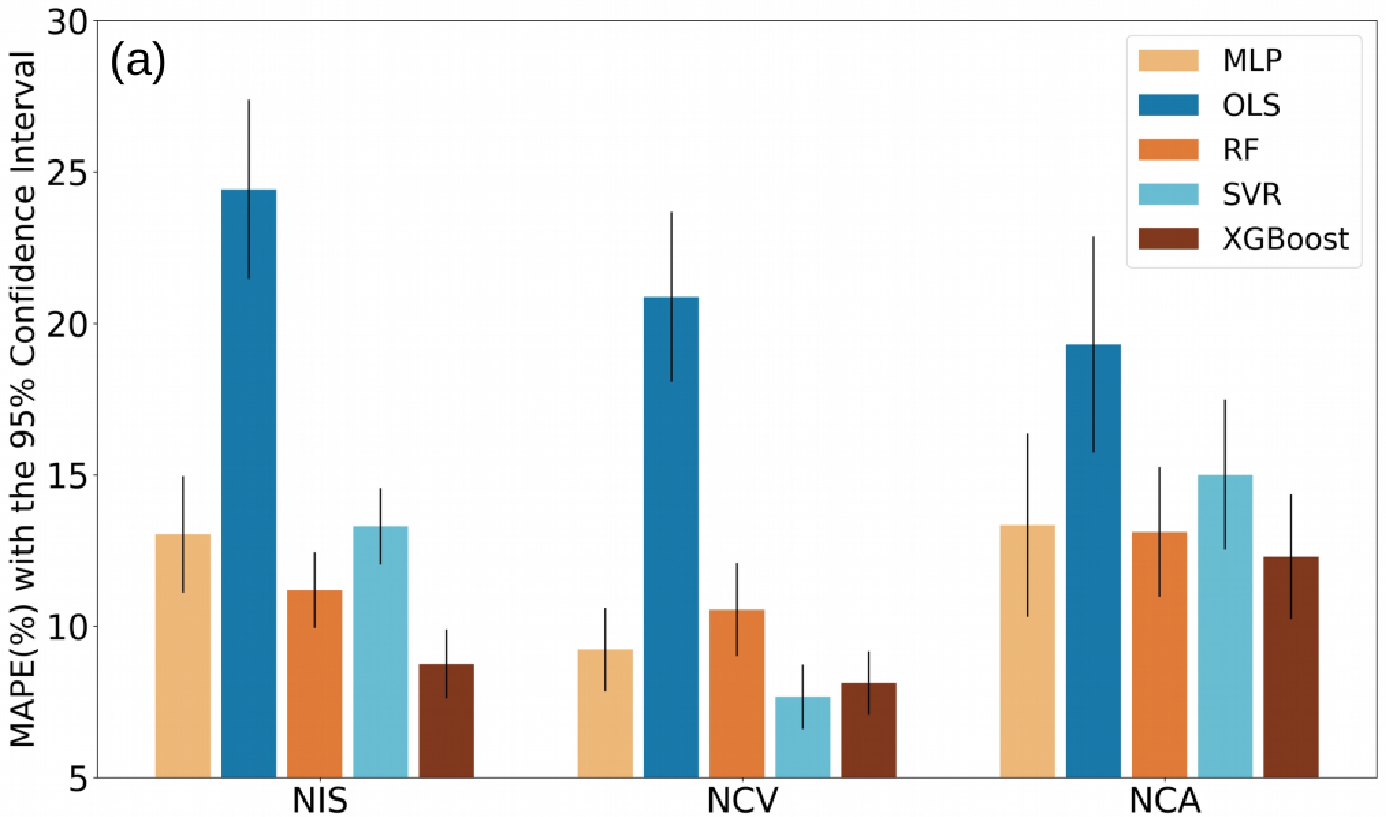}
        \end{minipage}
    \begin{minipage}{0.49\textwidth}
        \centering
		\label{fig:tx2_bar_error}
    \includegraphics[width=1.0\textwidth, height=0.25\textheight]{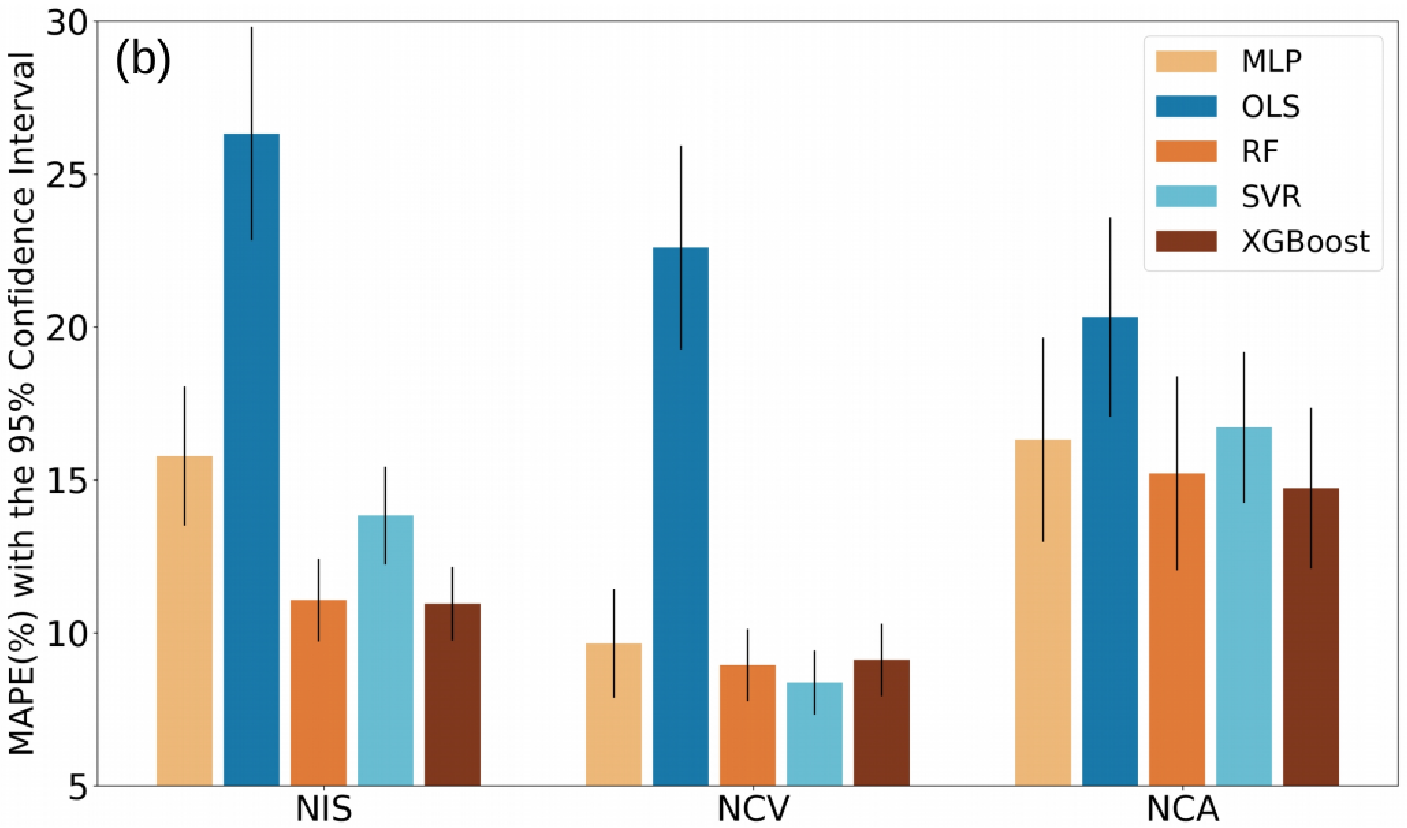}
        \end{minipage}
    \caption{Mean Absolute Percentage Error (MAPE) for the Nvidia AGX \textbf{(a)} and TX2 \textbf{(b)} \textsc{gpu} platforms with the corresponding 95\% Confidence Interval. In this figure we compare the 5 prediction models for exploring: New  Image  Sizes  (NIS),  New \textsc{cnn} Variants (NCV) and New \textsc{cnn} Architectures (NCA).}
    \label{fig:New_MAPE}
    \end{figure*} 

\begin{figure*}[!ht]
\begin{minipage}[c]{.49\linewidth}
      \includegraphics[width=1\textwidth, height=0.15\textheight]{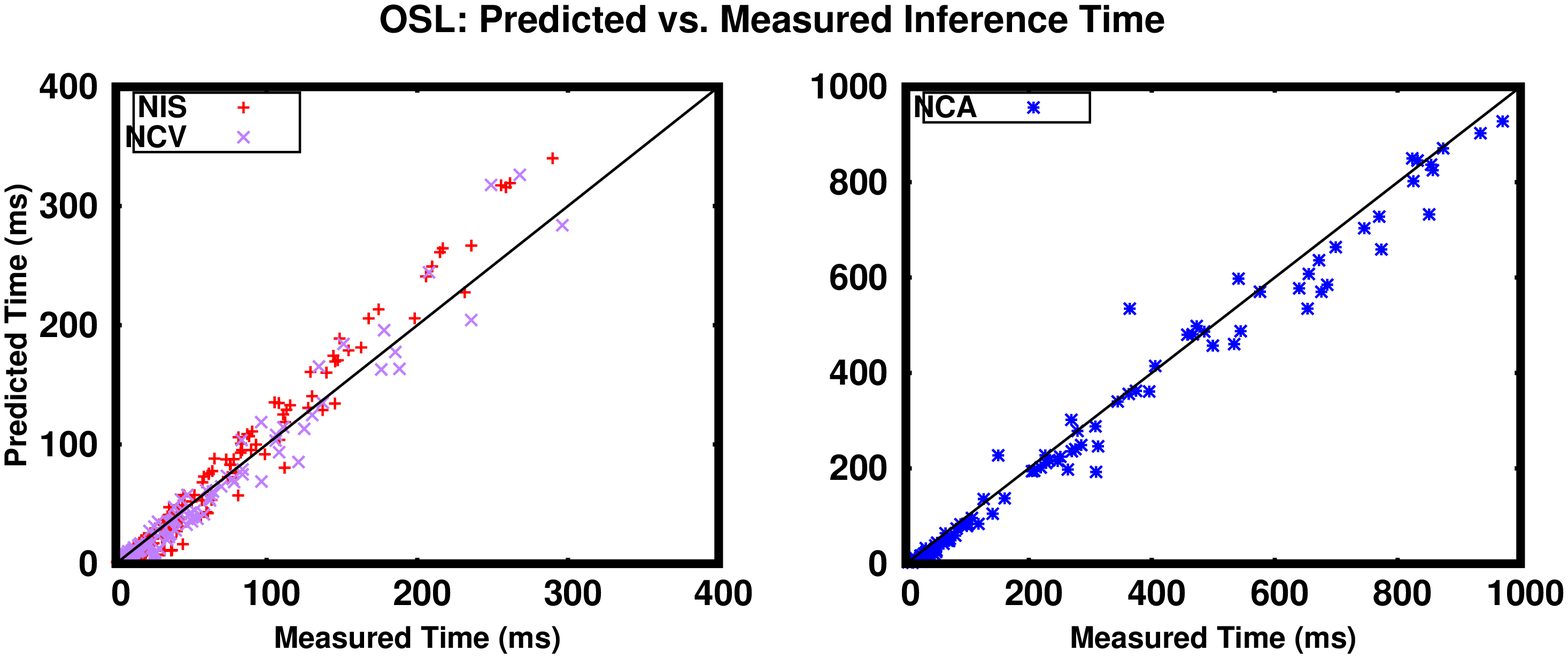}
      
      \caption{\small Predicted vs. Measured Inference Time for OLS (AGX). } 
      \label{fig:OSL}
   \end{minipage} \hfill
   \begin{minipage}[c]{.49\linewidth}
      \includegraphics[width=1\textwidth, height=0.15\textheight]{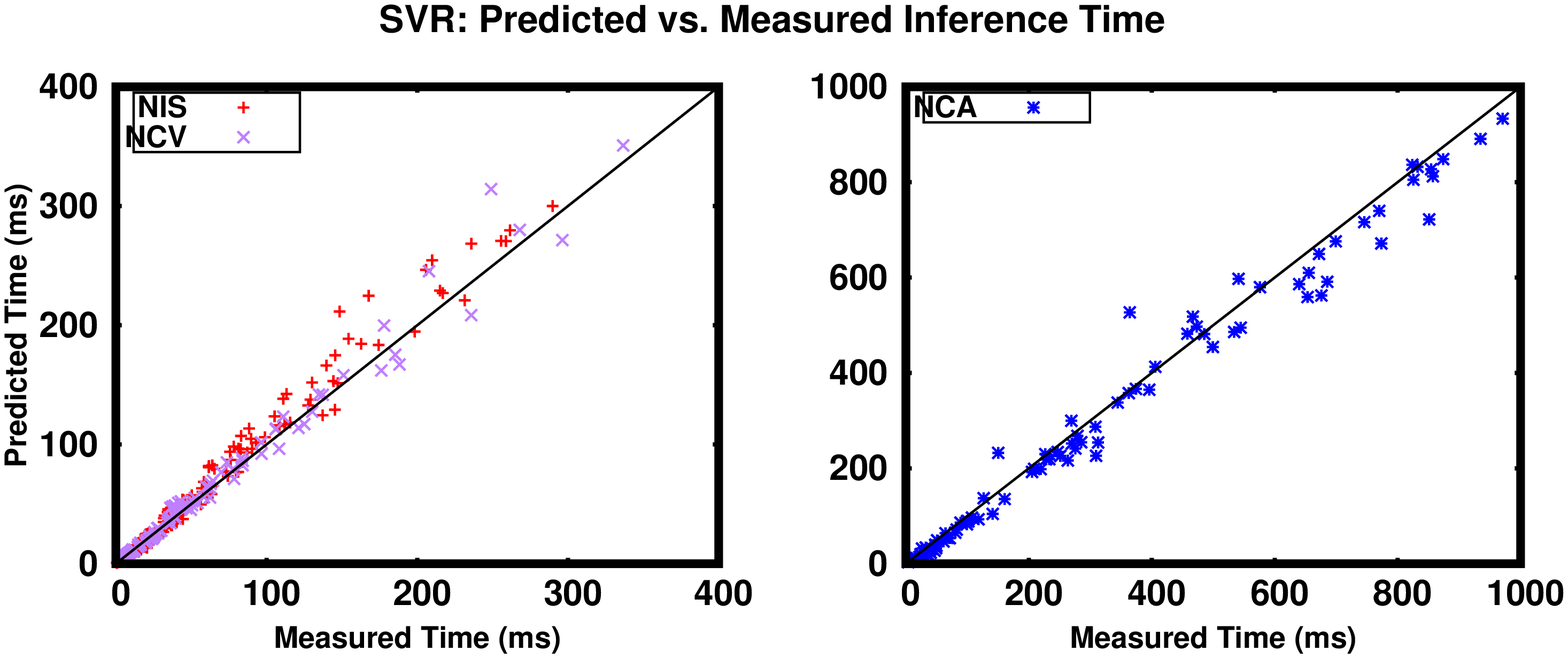}
      
      \caption{\small Predicted vs. Measured Inference Time for SVR (AGX). } 
      \label{fig:SVR}
   \end{minipage}
\end{figure*}

\begin{figure*}[!ht]
	\begin{minipage}[c]{.49\linewidth}
      	\includegraphics[width=1\textwidth, height=0.15\textheight]{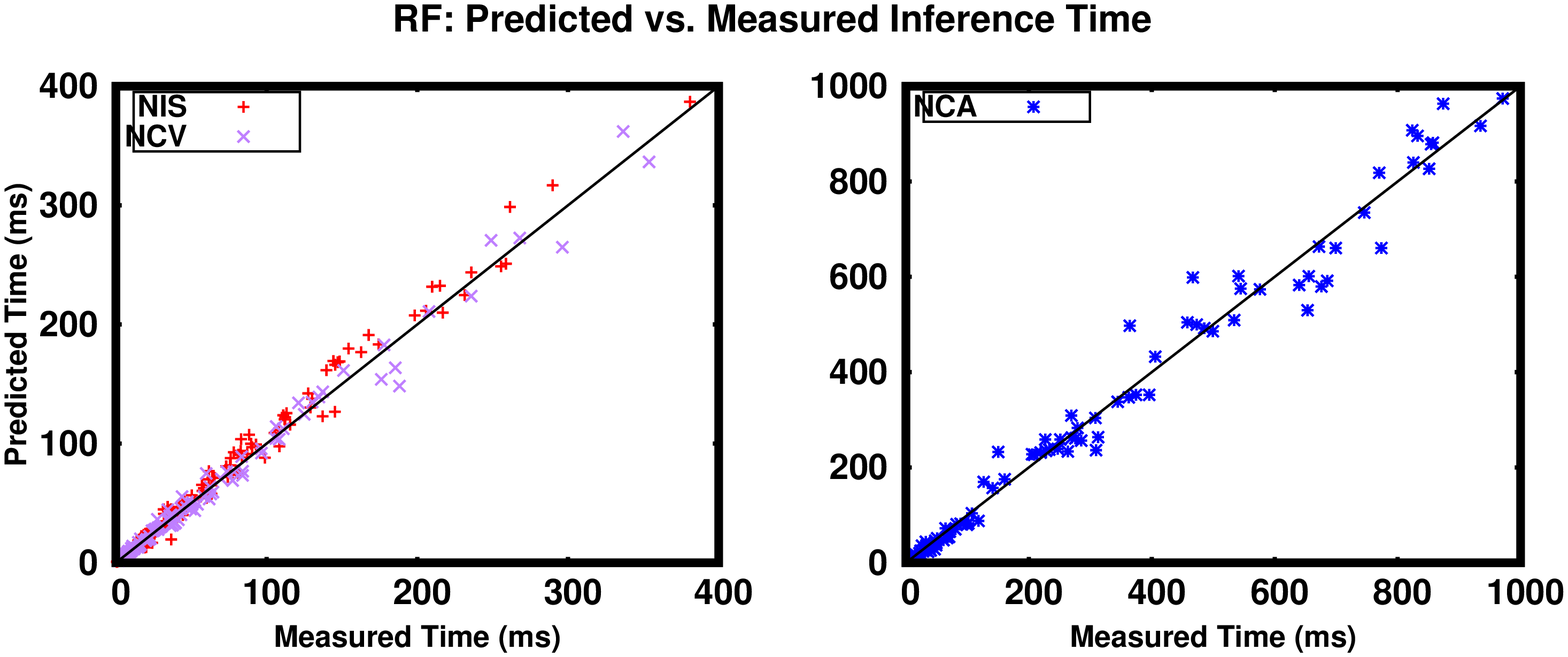}
      	      	
      	\caption{\small Predicted vs. Measured Inference Time for RF (AGX). }
      	\label{fig:RF}
  	\end{minipage} \hfill
   	\begin{minipage}[c]{.49\linewidth}
      \includegraphics[width=1\textwidth, height=0.15\textheight]{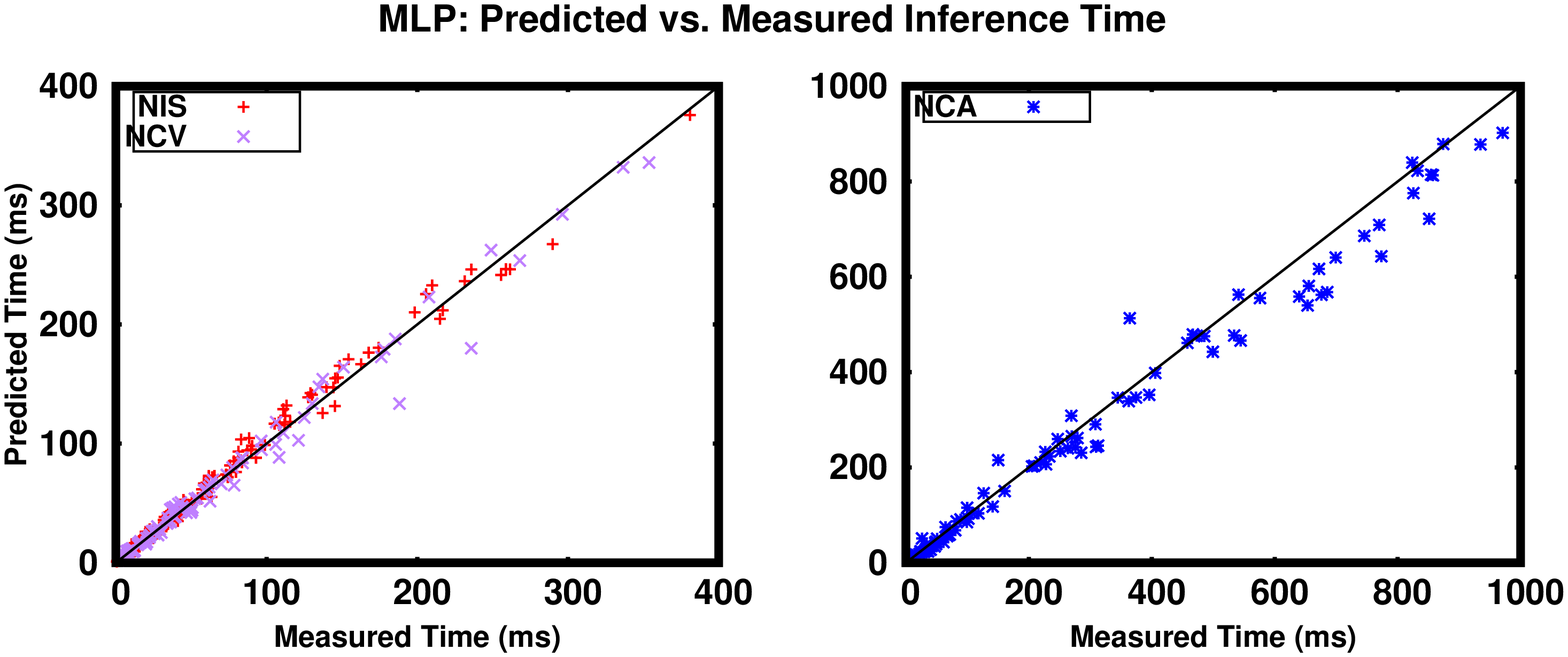}      	
      	\caption{\small Predicted vs. Measured Inference Time for MLP (AGX). }
\label{fig:MLP}   
   \end{minipage}
\end{figure*}

\begin{figure}[!ht]
    \centering
    \includegraphics[width=0.5\textwidth, height=0.15\textheight] {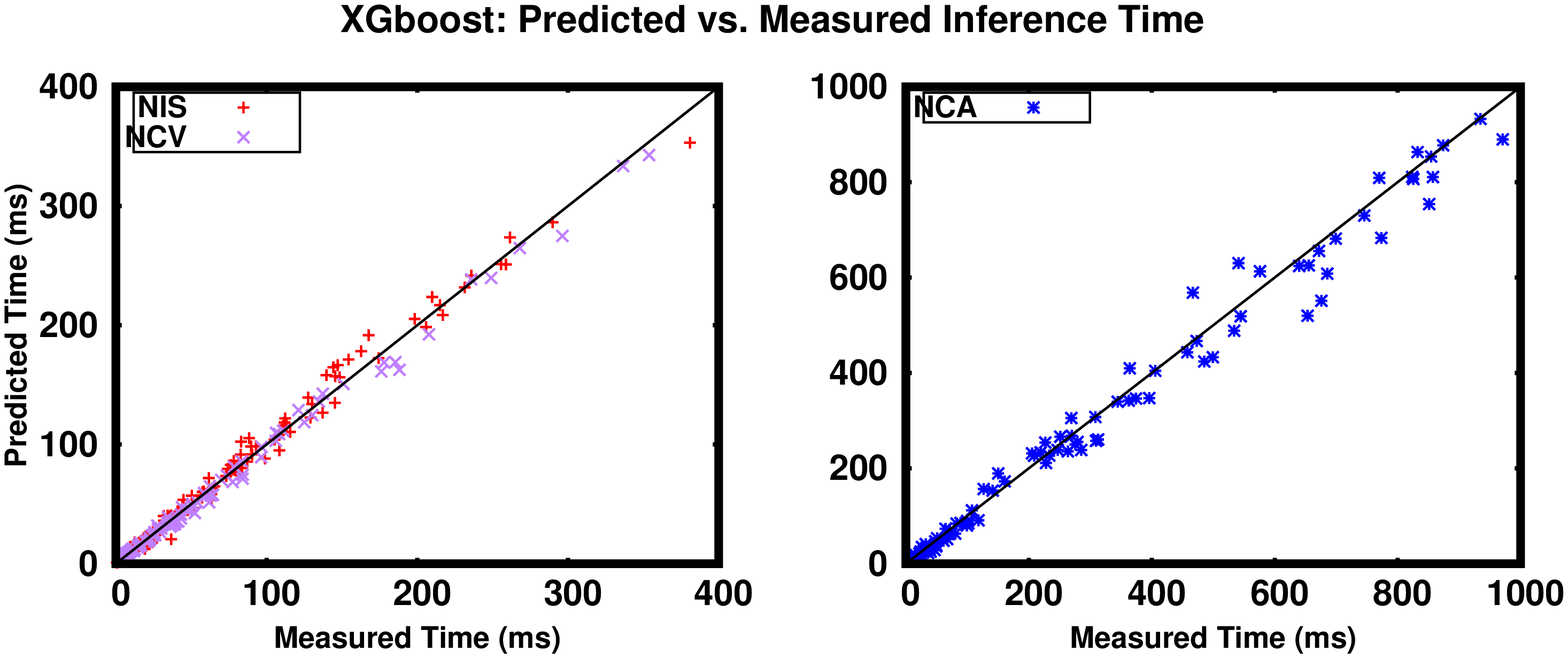}
    \caption{\small Predicted vs. Measured Inference Time for XGBoost (AGX). }
    \label{fig:XGBOOST}
\end{figure}

\begin{table*}[!ht]
\captionsetup{justification=centering}
    \captionof{table}{\textsc{Prediction models analysis: Accuracy, training, tuning, and latency}}
\fontsize{8}{8}\selectfont
\makebox[\linewidth]{
\label{tab:final_table}
\begin{tabular}{|c|c|c|c|c|c|c|c|c|c|c|l|c|l|c|l|}
\hline
\multicolumn{2}{|c|}{} &
  \multicolumn{2}{c|}{\textbf{MAPE}} &
  \multicolumn{2}{c|}{Training time} &
  \multicolumn{2}{c|}{\begin{tabular}[c]{@{}c@{}}Grid Search \\ execution time\end{tabular}} &
  \multicolumn{2}{c|}{\begin{tabular}[c]{@{}c@{}}Total number of \\ hyperparameters\end{tabular}} &
  \multicolumn{6}{c|}{Prediction latency} \\ \hline
\begin{tabular}[c]{@{}c@{}}Prediction\\ Model\end{tabular} &
  \begin{tabular}[c]{@{}c@{}}Test\\ Data\end{tabular} &
  AGX &
  TX2 &
  AGX &
  TX2 &
  AGX &
  TX2 &
  AGX &
  TX2 &
  \multicolumn{2}{c|}{AGX} &
  \multicolumn{2}{c|}{TX2} &
  \multicolumn{2}{c|}{\begin{tabular}[c]{@{}c@{}}Intel\\ Xeon\end{tabular}} \\ \hline
\multirow{3}{*}{\textbf{OLS}} &
  NIS &
  24.43\% &
  26.32\% &
  \multirow{3}{*}{\textbf{8.14 us}} &
  \multirow{3}{*}{\textbf{7.3 us}} &
  \multicolumn{2}{c|}{\multirow{3}{*}{0}} &
  \multicolumn{2}{c|}{\multirow{3}{*}{0}} &
  \multicolumn{2}{c|}{\multirow{3}{*}{319.6 ns}} &
  \multicolumn{2}{c|}{\multirow{3}{*}{468.42 ns}} &
  \multicolumn{2}{c|}{\multirow{3}{*}{113.93 ns}} \\ \cline{2-4}
 &
  NCV &
  20.87\% &
  22.95\% &
   &
   &
  \multicolumn{2}{c|}{} &
  \multicolumn{2}{c|}{} &
  \multicolumn{2}{c|}{} &
  \multicolumn{2}{c|}{} &
  \multicolumn{2}{c|}{} \\ \cline{2-4}
 &
  NCA &
  19.30\% &
  20.32\% &
   &
   &
  \multicolumn{2}{c|}{} &
  \multicolumn{2}{c|}{} &
  \multicolumn{2}{c|}{} &
  \multicolumn{2}{c|}{} &
  \multicolumn{2}{c|}{} \\ \hline
\multirow{3}{*}{\textbf{MLP}} &
  NIS &
  13.04\% &
  15.78\% &
  \multirow{3}{*}{1.53 s} &
  \multirow{3}{*}{722 ms} &
  \multirow{3}{*}{10.35 hr} &
  \multirow{3}{*}{11.82 hr} &
  \multirow{3}{*}{14400} &
  \multirow{3}{*}{15840} &
  \multicolumn{2}{c|}{\multirow{3}{*}{22.65 us}} &
  \multicolumn{2}{c|}{\multirow{3}{*}{30.90 us}} &
  \multicolumn{2}{c|}{\multirow{3}{*}{9.95 us}} \\ \cline{2-4}
 &
  NCV &
  9.23\% &
  9.65\% &
   &
   &
   &
   &
   &
   &
  \multicolumn{2}{c|}{} &
  \multicolumn{2}{c|}{} &
  \multicolumn{2}{c|}{} \\ \cline{2-4}
 &
  NCA &
  13.34\% &
  16.31\% &
   &
   &
   &
   &
   &
   &
  \multicolumn{2}{c|}{} &
  \multicolumn{2}{c|}{} &
  \multicolumn{2}{c|}{} \\ \hline
\multirow{3}{*}{\textbf{SVR}} &
  NIS &
  13.30\% &
  13.84\% &
  \multirow{3}{*}{127 ms} &
  \multirow{3}{*}{191 ms} &
  \multirow{3}{*}{23.61 hr} &
  \multirow{3}{*}{21.43 hr} &
  \multirow{3}{*}{18144} &
  \multirow{3}{*}{18144} &
  \multicolumn{2}{c|}{\multirow{3}{*}{41.79 us}} &
  \multicolumn{2}{c|}{\multirow{3}{*}{52.14 us}} &
  \multicolumn{2}{c|}{\multirow{3}{*}{20.76 us}} \\ \cline{2-4}
 &
  NCV &
  \textbf{7.67\%} &
  \textbf{8.37\%} &
   &
   &
   &
   &
   &
   &
  \multicolumn{2}{c|}{} &
  \multicolumn{2}{c|}{} &
  \multicolumn{2}{c|}{} \\ \cline{2-4}
 &
  NCA &
  15.00\% &
  16.72\% &
   &
   &
   &
   &
   &
   &
  \multicolumn{2}{c|}{} &
  \multicolumn{2}{c|}{} &
  \multicolumn{2}{c|}{} \\ \hline
\multirow{3}{*}{\textbf{RF}} &
  NIS &
  11.19\% &
  11.07\% &
  \multirow{3}{*}{4.93 s} &
  \multirow{3}{*}{2.01 s} &
  \multirow{3}{*}{4.69 hr} &
  \multirow{3}{*}{4.22 hr} &
  \multirow{3}{*}{71280} &
  \multirow{3}{*}{71280} &
  \multicolumn{2}{c|}{\multirow{3}{*}{1.03 ms}} &
  \multicolumn{2}{c|}{\multirow{3}{*}{1.45 ms}} &
  \multicolumn{2}{c|}{\multirow{3}{*}{393.3 us}} \\ \cline{2-4}
 &
  NCV &
  10.55\% &
  \textbf{8.95\%} &
   &
   &
   &
   &
   &
   &
  \multicolumn{2}{c|}{} &
  \multicolumn{2}{c|}{} &
  \multicolumn{2}{c|}{} \\ \cline{2-4}
 &
  NCA &
  13.11\% &
  15.22\% &
   &
   &
   &
   &
   &
   &
  \multicolumn{2}{c|}{} &
  \multicolumn{2}{c|}{} &
  \multicolumn{2}{c|}{} \\ \hline
\multirow{3}{*}{\textbf{XGBoost}} &
  NIS &
  \textbf{8.75\%} &
  \textbf{10.95\%} &
  \multirow{3}{*}{163 ms} &
  \multirow{3}{*}{914 ms} &
  \multirow{3}{*}{\textbf{11.27 mn}} &
  \multirow{3}{*}{\textbf{15.17 mn}} &
  \multirow{3}{*}{\textbf{237}} &
  \multirow{3}{*}{\textbf{238}} &
  \multicolumn{2}{c|}{\multirow{3}{*}{2.03 us}} &
  \multicolumn{2}{c|}{\multirow{3}{*}{3.49 us}} &
  \multicolumn{2}{c|}{\multirow{3}{*}{175.20 ns}} \\ \cline{2-4}
 &
  NCV &
  \textbf{8.12\%} &
  \textbf{9.11\%} &
   &
   &
   &
   &
   &
   &
  \multicolumn{2}{c|}{} &
  \multicolumn{2}{c|}{} &
  \multicolumn{2}{c|}{} \\ \cline{2-4}
 &
  NCA &
  \textbf{12.29\%} &
  \textbf{14.73\%} &
   &
   &
   &
   &
   &
   &
  \multicolumn{2}{c|}{} &
  \multicolumn{2}{c|}{} &
  \multicolumn{2}{c|}{} \\ \hline
\end{tabular}
}
\end{table*}

\subsection{Results and Discussion}
After detailing the evaluation methodology, this section presents and discusses the obtained modeling results in three perspectives:
First, we discuss the predicted \textsc{cnn} inference times using the 5 ML prediction models compared to the measured ones.
Second, the Mean Absolute Percentage Error (MAPE) is presented to evaluate the accuracy of each ML model. We also discuss the hyperparameters configuration.
In the third part of this section, we compare execution times for training and for tuning, the number of tested hyperparameters' configurations and finally the prediction latency which corresponds to the execution time of the 5 prediction approaches on
3 different Hardware platforms: on the Jeston AGX \textsc{gpu}, on the Jetson TX2 \textsc{gpu} and on a development station Intel Xeon.
Implementing our predictors on edge \textsc{gpu} platforms will allow to adapt the executed \textsc{cnn} to application constraints at run time. 

\subsubsection{Predicted vs Measured \textsc{cnn} inference time}
Figures \ref{fig:OSL}, \ref{fig:SVR}, \ref{fig:RF}, \ref{fig:MLP}, and \ref{fig:XGBOOST} present the comparison between the predicted (y axis) and the measured (x axis) \textsc{cnn} inference time on the AGX platform. 
The comparison has been realized for NIS, NCV (on the left of the figures) and NCA (on the right of the figures). The used dataset has been detailed in section \ref{sub:Evaluation_Methodology} and Table \ref{tab:parameters}.

For NIS and NCV, the predicted execution times are very close to the measured values, in all of the prediction  models except for OLS (see figure \ref{fig:OSL}). 
As we have trained our prediction models on different \textsc{cnn} variants and different input image sizes, the prediction models have been able to capture the correlation between different \textsc{cnn} variants and input sizes. We can also notice that OLS overestimates the \textsc{cnn} inference time especially for high execution time values. 
This is due to the non-linearity between input features and the \textsc{cnn} inference time where the computation complexity and memory requirements are very high. This result confirms also that the inference times could not be accurately estimated by using a linear regression .

For NCA, the difference between the predicted and the measured execution times are higher than for NIS and NCV, in particular for MLP, OLS and SVR prediction models. 
As explained in the previous section, the reason behind this drawback is the fact that in NCA, new  and randomly generated \textsc{cnn}s are explored. 
For these reasons, the predictions models are less accurate for these unexplored and unseen \textsc{cnn} architectures.

In the appendix of the paper, figures \ref{fig:OSL-tx2}, \ref{fig:SVR-tx2}, \ref{fig:RF-tx2}, \ref{fig:MLP-tx2}, and \ref{fig:XGBOOST-tx2} show the results for the NVIDIA Jetson TX2 platform. We observe the same conclusions as for the NVIDIA AGX.
 
\subsubsection{Prediction accuracy and hyperparameters configuration analysis }
To evaluate the accuracy of our prediction models, the Mean Absolute Percentage Error (MAPE) has been chosen. 
Figure \ref{fig:New_MAPE}  and Table \ref{tab:final_table} give the obtained MAPE for the 5 studied prediction models for NIS, NCV and NCA.
The prediction models use data obtained by profiling the benchmarks (Table \ref{tab:parameters}) on two edge \textsc{gpu} platforms (see section \ref{subsub:benchmark}).

From figure \ref{fig:New_MAPE} and Table \ref{tab:final_table}, we can note that the MAPE average values are mostly between $\sim 7\%$ in the best case and $\sim 26\%$ in the worst case. 
We can also notice that the lowest MAPE values have been obtained for NIS and NCV. 

In general we can say that for NIS, NCV and NCA, XGBoost outperforms the other prediction models and offers the lowest MAPE values. XGBoost is among the most powerful ML algorithms. The Boosting technique, to enhance the prediction accuracy through many estimators arranged sequentially, makes it very accurate. However, we noticed that XGBoost is sensitive to hyperparameters values. These values need to be set carefully in order to achieve the best performances. 

RF is composed of different decision trees trained on random subsets of training data samples and features. This property  helps to reduce the variance error. Nevertheless, prediction approaches based on ensemble learning, such as RF, need to be trained on different scales of data in order to obtain an accurate mapping of features and execution times. 
In addition, by increasing the number or the depth of the decision trees, the prediction accuracy converges according to our the experiments.
These 2 factors increase the complexity of the training and the latency of the prediction in RF. For this reason, when a short interval of time for tuning and training is desired, XGBoost will be more efficient than RF.

MLP has generally good performances with a slight loss of generalization for NIS and NCA. This can be due to its nature that tends to overfit data. Furthermore, MLP is very sensitive to the variation of hyperparameters, which makes it very hard to tune. The size of the MLP network has an important impact on the accuracy. We noticed that large MLP networks are prone to overfitting compared to the small ones.

SVR is less accurate for NIS and NCA compared to the aforementioned models.  The variation in the MAPE values in the three exploration spaces is quite high which can be interpreted as overfitting. In terms of hyperparameters configuration, SVR is sensitive to the type of kernels and the cost (C). According to our results, linear kernels perform the best, whereas the very small values of C, lower than 1, lead to a huge loss of generalization.

As expected, OLS has the highest MAPE values because of its limited capacity to capture the different patterns in the training data. Moreover, the inclusion of non relevant features in OLS can drastically decrease the model’s accuracy. In this case, techniques for regularization such as LASSO and Ridge are recommended to improve the model performances. Due of time limitation, this point is considered as a possible extension.

If we compare the results of the two \textsc{gpu} platforms, Nvidia AGX and TX2, we notice that the MAPE is very similar.
This means that our modeling approach is easily adaptable to other platforms in order to obtain execution time prediction.\\

\subsubsection{Comparison of the 5 models in terms of training time, tuning cost and latency}
Table \ref{tab:final_table} summarizes the obtained statistics of the 5 tested modeling approaches.

In addition to MAPE average values, table \ref{tab:final_table} gives the training and tuning time of each prediction model for AGX and TX2 in Google Colaboratory\cite{bisong2019google}. 
We can observe, that except for SVR and XGBoost, the prediction models require more time to be trained for AGX than for TX2. 
This is due to the fact that the higher resource capacity of AGX allows to run more \textsc{cnn}s which gives a largest training data compared to TX2. Table \ref{tab:final_table} shows also that tuning the model hyperparameters is time consuming task, especially for models with high number of hyperparameters.

Tested hyperparameters values in grid search for the AGX and for the TX2 are different. This is due to the fact that \textsc{cnn} execution times ranges are different.
For the MLP and XGBoost, as both of them use the gradient descent algorithm, the training convergence for AGX and for TX2 is different.

To quantify the execution time of the 5 models, they have been executed on an  AGX, TX2 and Intel Xeon based Processor development station. Results are given in the three rightmost columns in Table \ref{tab:final_table}. The comparison of the execution latency is important to measure the performance of the 5 models to explore a large number of \textsc{cnn} architectures in a limited interval of time. In addition, when an online tuning of the \textsc{cnn} or the Hardware platform is needed, having a short execution time of the prediction model is interesting.
We note that in terms of speed, OLS outperforms all of the other models.
This is explained by the simplicity of OLS which is a simple linear equation. However OLS is the less accurate. XGBoost has a  smaller execution time compared to MLP, SVR and RF. Which makes this technique very interesting due to the good trade-off in terms of accuracy and latency.
The highest latency of RF is due to the complexity of exploring the decision trees included in the Random Forest.

\section{Conclusion}
\label{conclusion}
Edge computing is one of the area targeted by \textsc{cnn}-based applications.
Both \textsc{cnn}s and edge computing are highly growing and frequently changing domains. 
Finding the best matching between \textsc{cnn} model architecture and Hardware of the edge device under real time constraints is a very time consuming and complex task.
In this paper, we compared five state of the art ML-based models for \textsc{cnn} inference time prediction on edge \textsc{gpu}s. These 5 models are useful for an early performance estimation of \textsc{cnn}-based applications. Using these models, rapid design space exploration will be possible to guide the designer to an efficient \textsc{cnn} model and/or to the most adequate Hardware platform. 

We demonstrated that XGBoost and RF gave good execution time predictions, with an average accuracy  close to $92\%$ for NIS and NCV and $86\%$ for NCA. In terms of trade-off between: accuracy, model tuning complexity and prediction latency, XGBoost is the best approach. Exploring new synthetic \textsc{cnn}s, i.e called NCA in our paper, is more complex and less accurate than the other two spaces: NIS and NCV. 

As future work, we plan to extend our comparison to explore additional  \textsc{cnn}s' characteristics that may improve our prediction models. We also plan to extend our ML-based models to predict not only \textsc{cnn}s inference execution time but also energy consumption and resource utilization. Finally, we will also integrate the hardware configuration as input in our models to perform cross platform predictions. 

\bibliographystyle{IEEEtran}
\bibliography{ref.bib}

\appendix
\subsection{Obtained results on Jeston TX2}
\label{apdx:tx2}
\begin{figure}[H]
      \includegraphics[width=0.48\textwidth, height=0.16\textheight]{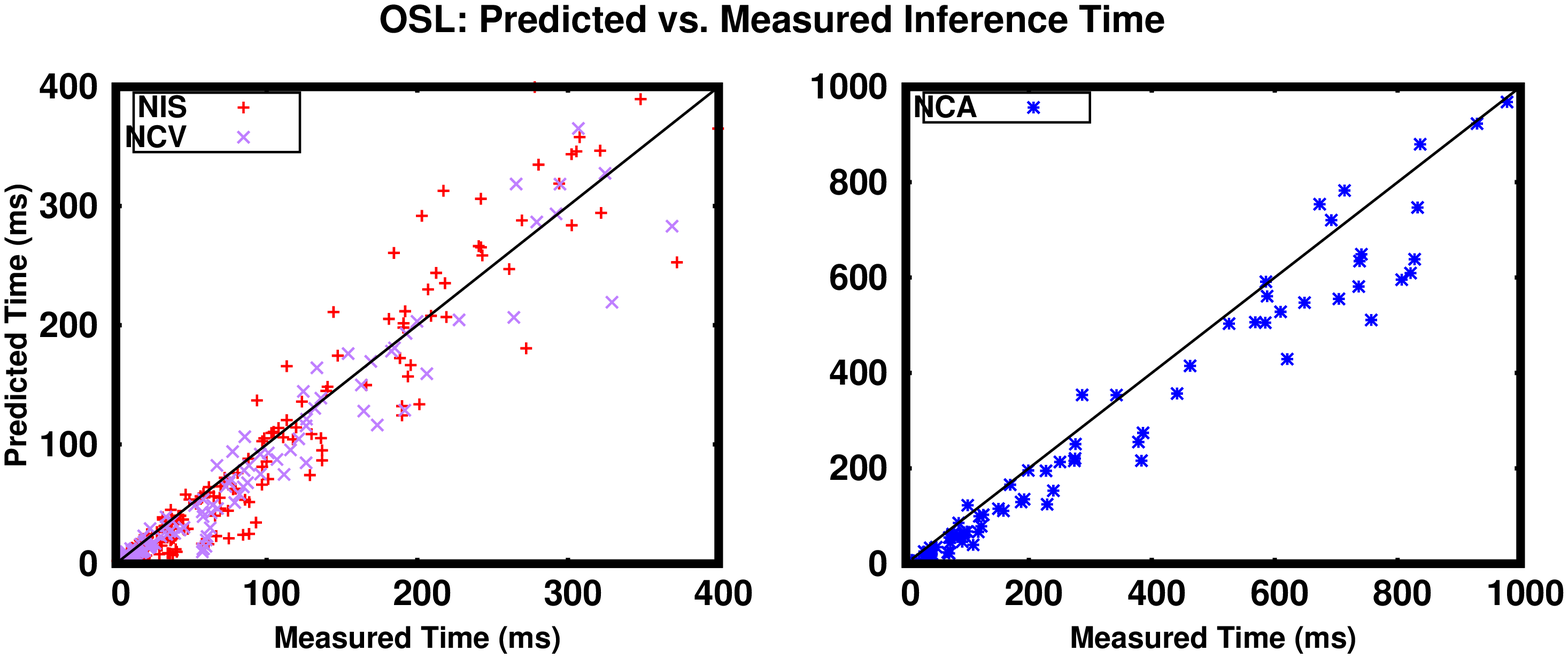}
      
      \caption{\small Predicted vs. Measured Inference Time for OLS (TX2).} 
      \label{fig:OSL-tx2}
\end{figure}
\begin{figure}[!ht]
\includegraphics[width=0.48\textwidth, height=0.18\textheight]{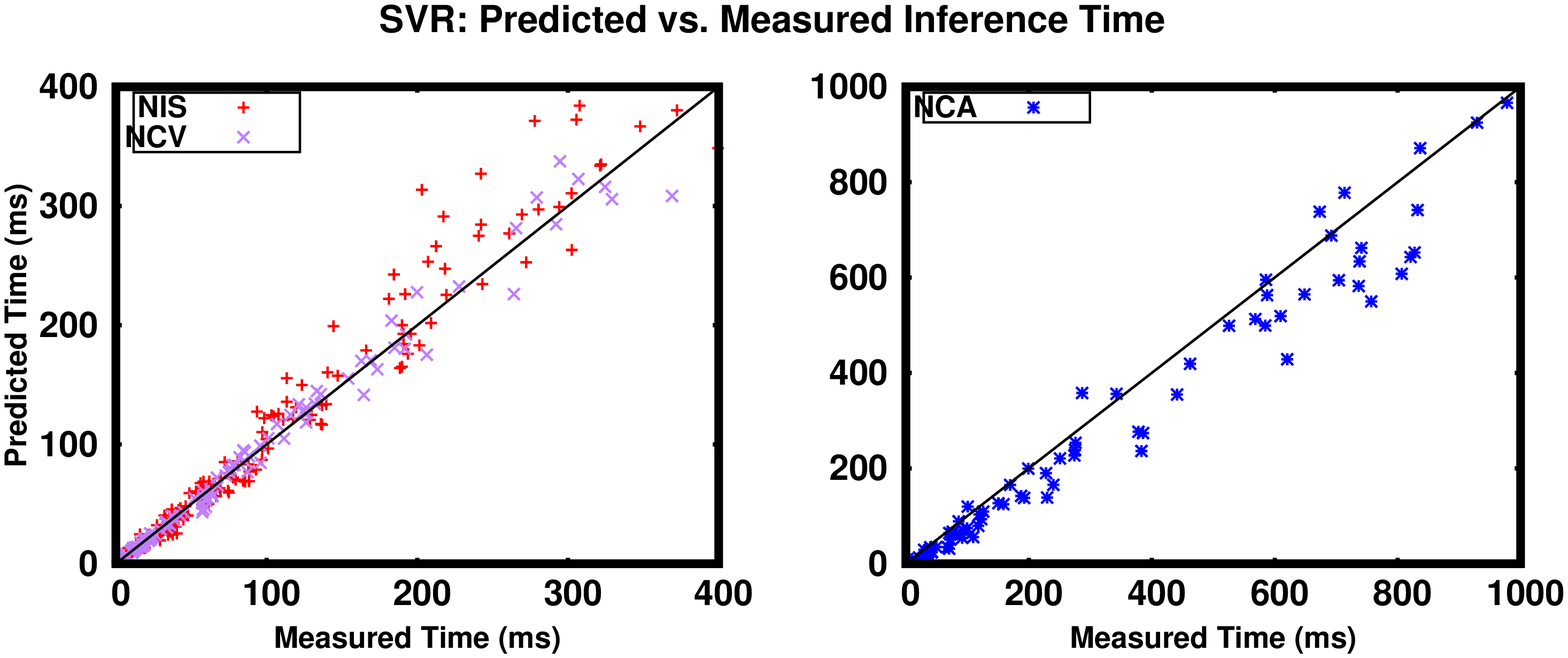}
      
      \caption{\small Predicted vs. Measured Inference Time for SVR (TX2). } 
      \label{fig:SVR-tx2}
\end{figure}

\begin{figure}[ht!]
      	\includegraphics[width=0.48\textwidth, height=0.16\textheight]{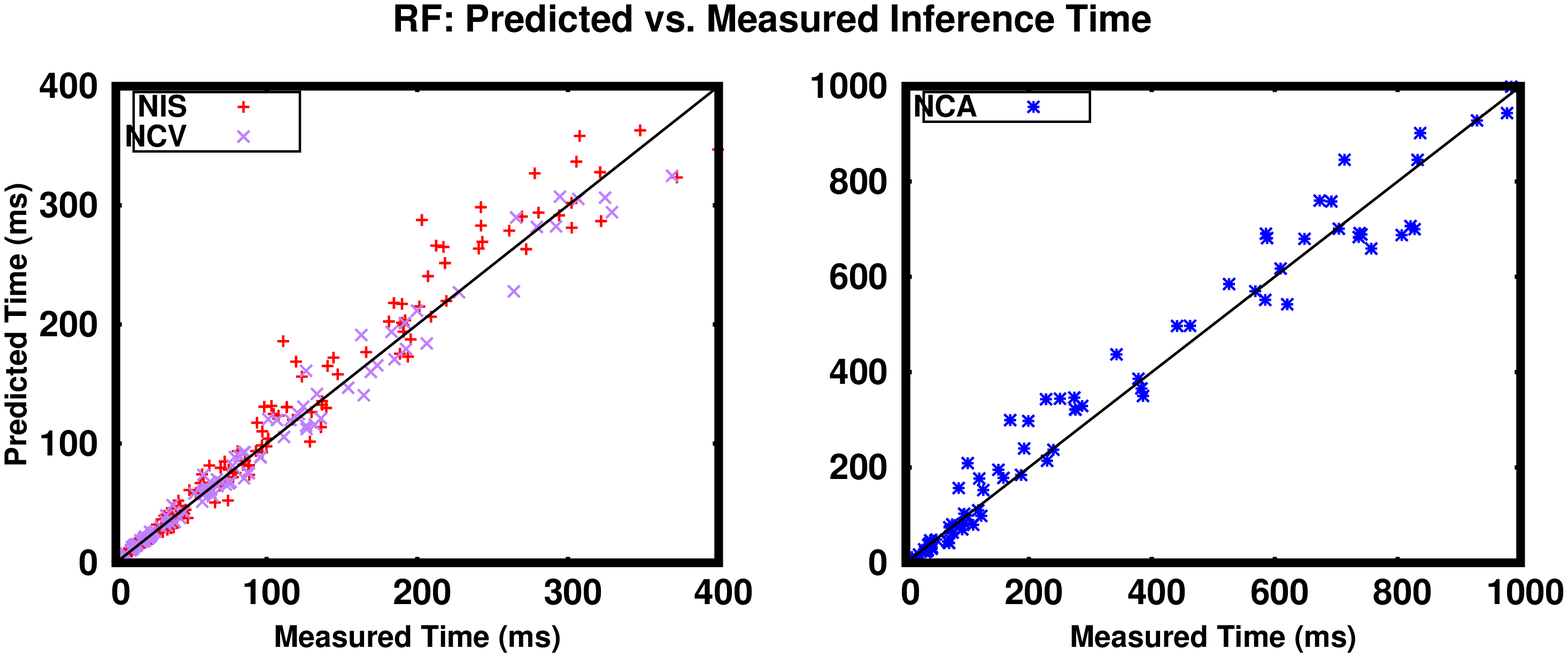}
      	      	
      	\caption{\small Predicted vs. Measured Inference Time for RF (TX2). }
      	\label{fig:RF-tx2}
\end{figure}
\begin{figure}[ht!]      
      \includegraphics[width=0.48\textwidth, height=0.16\textheight]{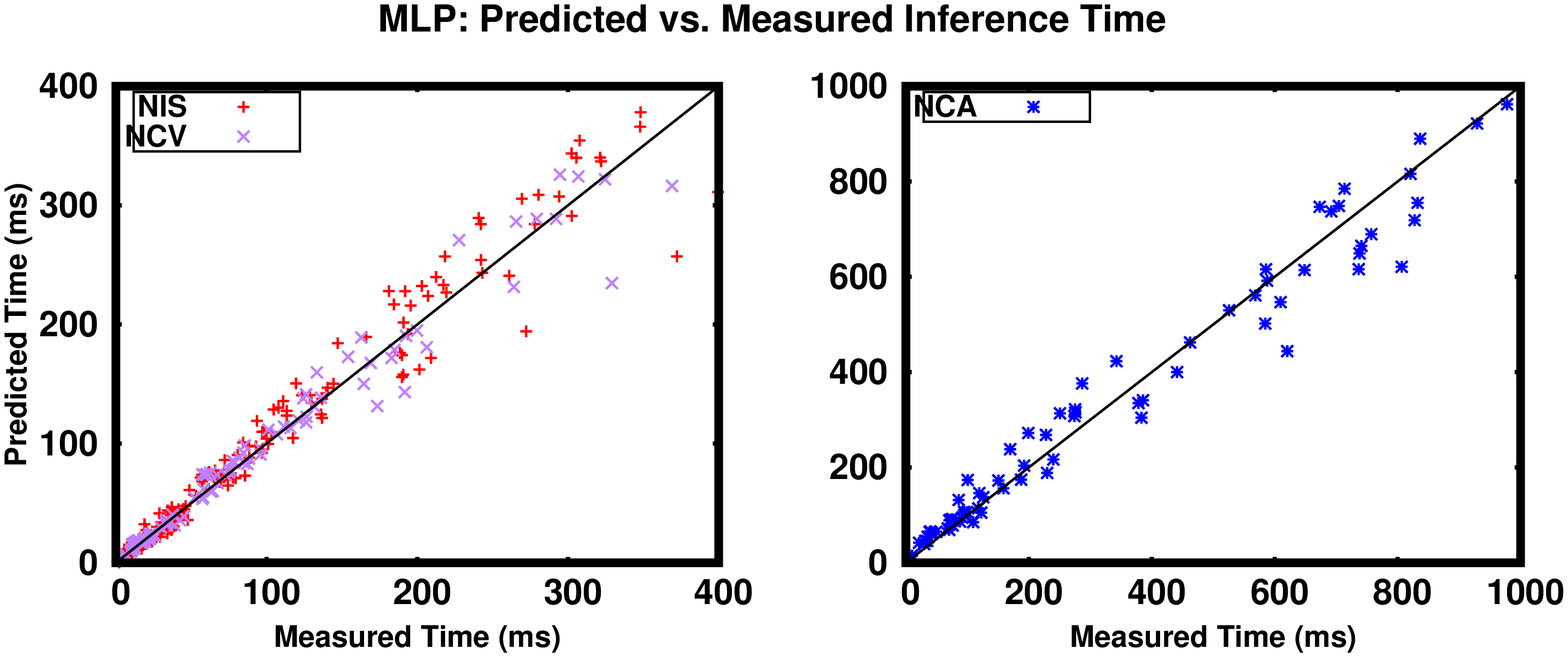}      	
      	\caption{\small Predicted vs. Measured Inference Time for MLP (TX2). }
\label{fig:MLP-tx2}   
\end{figure}

\begin{figure}[ht!]
    \includegraphics[width=0.48\textwidth, height=0.2\textwidth]{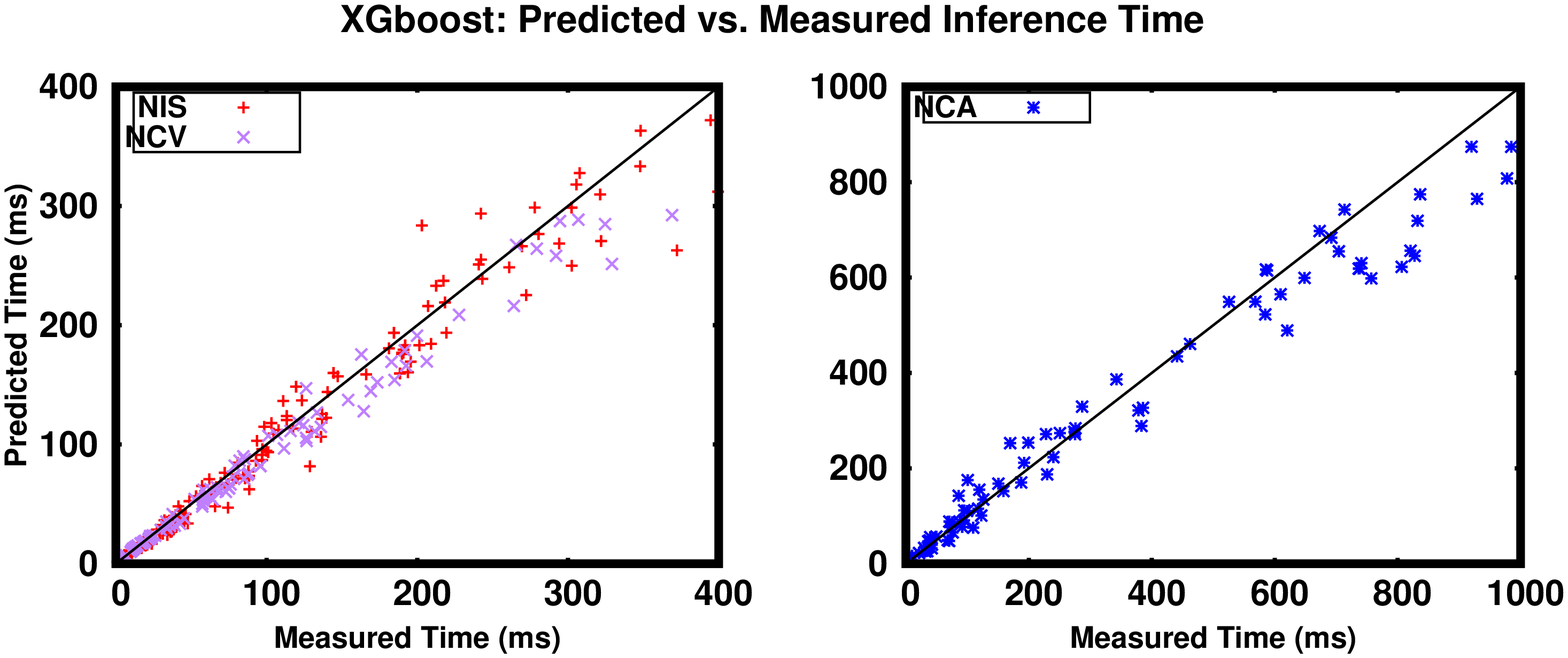}
    \caption{\small Predicted vs. Measured Inference Time for XGBoost (TX2). }
    \label{fig:XGBOOST-tx2}
\end{figure}

\end{document}